%% file: main.tex
\documentclass[sigconf]{acmart}

\usepackage{graphicx}
\usepackage{multirow}
\usepackage{threeparttable}
\usepackage{xspace}
\newcommand{\vpara}[1]{\vspace{0.04in}\noindent\textbf{#1}\xspace}
\usepackage{enumitem}
\usepackage{color}
\usepackage{colortbl}
\usepackage{bm}
\definecolor{Gray}{gray}{0.9}
\usepackage{booktabs,siunitx}
\usepackage{adjustbox}
\usepackage{makecell}
\usepackage[capitalise]{cleveref}

\usepackage{balance}
\usepackage{url}

\usepackage{tcolorbox}
\newtcolorbox{myprompt}[2][]
{
    colframe=black,      
    colback=gray!20,     
    boxrule=1pt,         
    arc=4pt,             
    left=10pt,           
    right=10pt,
    top=10pt,            
    bottom=10pt, 
    title=#2
}
\newtcolorbox{myprompt_double}[2][]
{
    colframe=black,      
    colback=gray!20,     
    boxrule=1pt,         
    arc=4pt,             
    left=10pt,           
    right=10pt,
    top=10pt,            
    bottom=10pt, 
    width=\textwidth,
    title=#2
}

\makeatletter
\def\@fnsymbol#1{%
  \ifcase#1\or
    \textdagger\or
    *\or
    \textdaggerdbl\or
    \S\or
    \P\or
    \|\or
    **\or
    \textdagger\textdagger\or
    \textdaggerdbl\textdaggerdbl
  \else
    \@ctrerr
  \fi}
\makeatother
\AtBeginDocument{%
  }

\copyrightyear{2026}
\acmYear{2026}
\setcopyright{cc}
\setcctype{by}
\acmConference[KDD '26]{Proceedings of the 32nd ACM SIGKDD Conference on Knowledge Discovery and Data Mining V.2}{August 09--13, 2026}{Jeju Island, Republic of Korea}
\acmBooktitle{Proceedings of the 32nd ACM SIGKDD Conference on Knowledge Discovery and Data Mining V.2 (KDD '26), August 09--13, 2026, Jeju Island, Republic of Korea}
\acmDOI{10.1145/3770855.3818198}
\acmISBN{979-8-4007-2259-2/2026/08}

\begin{document}
\title{Hyper-KGGen: A Skill-Driven Knowledge Extractor for High-Quality Knowledge Hypergraph Generation}
%
\author{Rizhuo Huang}
\authornote{Both authors contributed equally to this research.}
\affiliation{%
  \institution{Xi'an Jiaotong University}
  \department{State Key Laboratory of Human-Machine Hybrid Augmented Intelligence, Institute of Artificial Intelligence and Robotics}
  \city{Xi'an}
  \country{China}
}
\email{rizhuo.huang@stu.xjtu.edu.cn}

\author{Yifan Feng}
\authornotemark[1]
\affiliation{%
  \institution{Tsinghua University}
  \city{Beijing}
  \country{China}
}
\email{evanfeng97@gmail.com}

\author{Rundong Xue}
\affiliation{%
  \institution{Xi'an Jiaotong University}
  \department{State Key Laboratory of Human-Machine Hybrid Augmented Intelligence, Institute of Artificial Intelligence and Robotics}
  \city{Xi'an}
  \country{China}
}
\email{xuerundong2002@gmail.com}

\author{Shihui Ying}
\affiliation{%
  \institution{Shanghai University}
  \city{Shanghai}
  \country{China}
}
\email{shying@shu.edu.cn}

\author{Jun-Hai Yong}
\affiliation{%
  \institution{Tsinghua University}
  \city{Beijing}
  \country{China}
}
\email{yongjh@tsinghua.edu.cn}

\author{Chuan Shi}
\affiliation{%
  \institution{Beijing University of Posts and Telecommunications}
  \city{Beijing}
  \country{China}
}
\email{shichuan@bupt.edu.cn}

\author{Shaoyi Du}
\affiliation{%
  \institution{Xi'an Jiaotong University}
  \department{State Key Laboratory of Human-Machine Hybrid Augmented Intelligence, Institute of Artificial Intelligence and Robotics}
  \city{Xi'an}
  \country{China}
}
\email{dushaoyi@xjtu.edu.cn}
\authornote{Corresponding Authors.}

\author{Yue Gao}
\affiliation{%
  \institution{Tsinghua University}
  \city{Beijing}
  \country{China}
}
\email{gaoyue@tsinghua.edu.cn}
\authornotemark[2]

\renewcommand{\shortauthors}{Rizhuo Huang et al.}

\begin{abstract}
Knowledge hypergraphs surpass traditional binary knowledge graphs by encapsulating complex $n$-ary atomic facts, providing a more comprehensive paradigm for semantic representation. 
However, constructing high-quality hypergraphs remains challenging due to the \textit{scenario gap}: generic extractors struggle to generalize across diverse domains with specific jargon, while existing methods often fail to balance structural skeletons with fine-grained details.
To bridge this gap, we propose \textbf{Hyper-KGGen}, a skill-driven framework that reformulates extraction as a dynamic skill-evolving process. 
First, Hyper-KGGen employs a \textit{coarse-to-fine} mechanism to systematically decompose documents, ensuring full-dimensional coverage from binary links to complex hyperedges. 
Crucially, it incorporates an \textit{adaptive skill acquisition} module that actively distills domain expertise into a Global Skill Library. This is achieved via a stability-based feedback loop, where extraction stability serves as a relative reward signal to induce high-quality skills from unstable traces and missed predictions.
Additionally, we present \textbf{HyperDocRED}, a rigorously annotated benchmark for document-level knowledge hypergraph extraction. 
Experiments demonstrate that Hyper-KGGen significantly outperforms strong baselines, validating that evolved skills provide substantially richer guidance than static few-shot examples in multi-scenario settings. The source codes are available at \url{https://github.com/Rizrock/Hyper-KGGen}.
\end{abstract}

\ccsdesc[500]{Computing methodologies~Knowledge representation and reasoning}

\keywords{Knowledge Extraction, Knowledge Hypergraphs, Data Generation}

\maketitle


\input{section/1_intro}
\input{section/2_rel}
\input{section/3_pre}
\input{section/4_method}

\input{section/5_exp}
\input{section/6_con}

\bibliographystyle{ACM-Reference-Format}
\bibliography{ref}

\input{section/7_appendix}

\end{document}

%% file: section/1_intro.tex
\section{Introduction}
\label{sec:intro}

Transforming unstructured text into structured knowledge representations is a fundamental pursuit in artificial intelligence~\cite{schneider1973course,angeli2015leveraging}, serving as the backbone for advanced reasoning and retrieval augmented generation (RAG) systems. Traditionally, Knowledge Graphs (KGs) have been the gold standard, modeling information as binary relations, formulated as \textit{(head, relation, tail)} triples. However, this formulation often functions merely as a skeletal representation, oversimplifying the rich semantic complexities of the real world. A substantial fraction of facts are inherently $n$-ary ($n > 2$), involving crucial contextual qualifiers such as time, location, and conditions~\cite{luo2025hypergraphrag}. Reducing these high-order dependencies into independent pairwise correlations inevitably leads to information loss and semantic ambiguity. Consequently, \textbf{Knowledge Hypergraphs}~\cite{fatemi2019knowledge, zeng2023multi}, which model atomic facts as unified hyperedges connecting multiple entities, have emerged as a superior paradigm for preserving the comprehensive fidelity of raw corpora~\cite{liu2020generalizing,rosso2020beyond,wen2016representation}.

Despite the representational advantages of hypergraphs, effective extraction remains an open challenge. Existing hypergraph extraction methods typically focus heavily on identifying complex high-order relations. While effective for specific patterns, they often exhibit weakness in capturing fundamental low-order pairwise associations, leading to a "top-heavy" structure that lacks foundational connectivity. This imbalance hinders modeling full-dimensional associative knowledge, where simple links and complex hyperedges must coexist to form a coherent semantic network.

\begin{figure}
    \centering
    \includegraphics[width=0.85\linewidth]{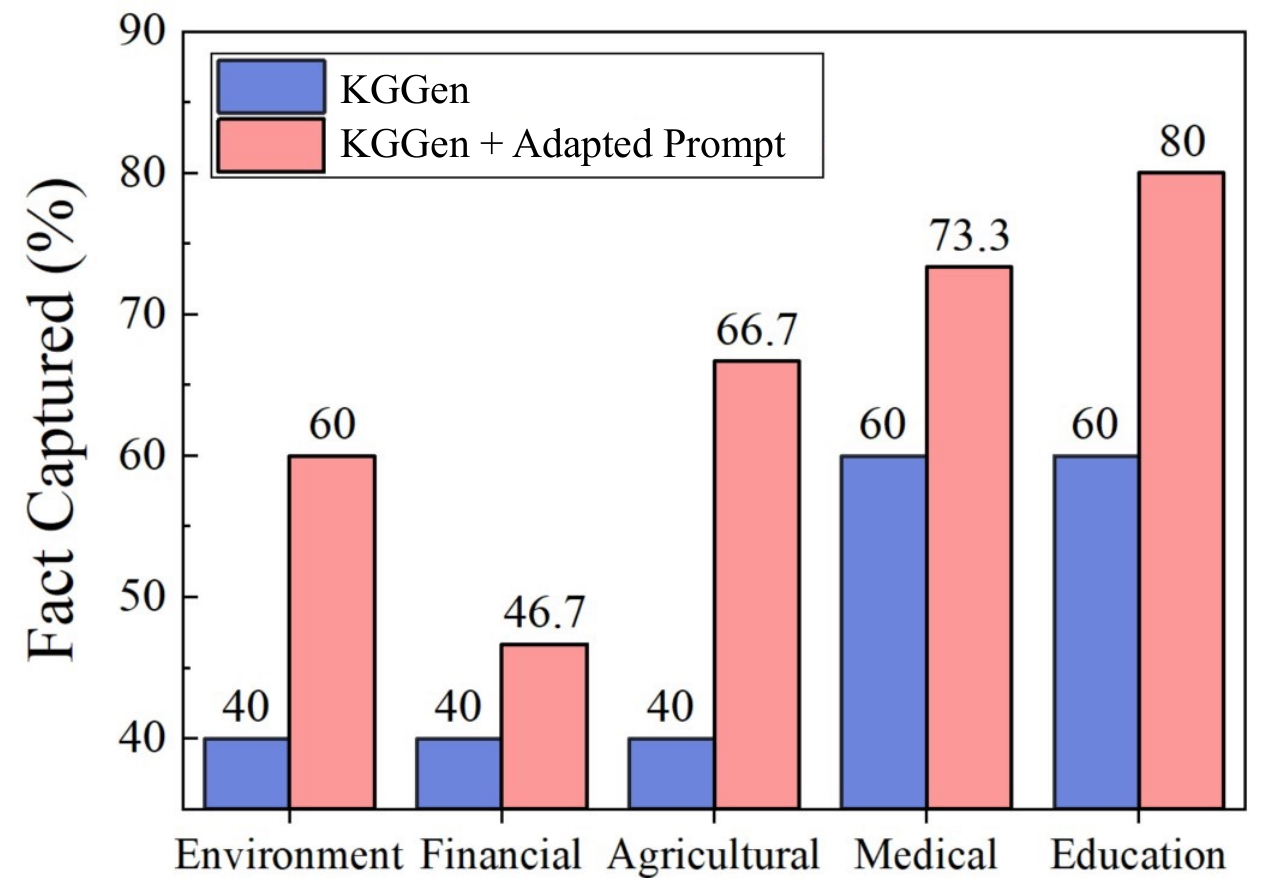}
    \caption{Illustration of the Scenario Gap of general prompting and domain-specific prompting. Using generic prompts in KGGen leads to suboptimal knowledge extraction. By contrast, applying domain-specific ``adaptive prompts'' substantially improves the model's ability to extract facts.}
    \label{fig:motivation}
\end{figure}

Furthermore, a significant barrier to high-quality extraction lies in the \textbf{Scenario Gap}. As illustrated in \cref{fig:motivation}, generic extraction models struggle to generalize across diverse domains (e.g., biomedical, legal, financial) due to the presence of industry jargon, implicit logic, and domain-specific boundaries. Our preliminary analysis reveals that while generic prompts yield suboptimal performance, manually optimizing prompts for specific domains significantly boosts extraction quality. This observation confirms that the model possesses the latent capability to handle domain tasks but lacks the explicit \textit{Skill} to align with scenario-specific constraints. However, manually crafting optimal prompts for every possible domain is unscalable. Therefore, there is an urgent need for an adaptive strategy that can automatically bridge this gap, transforming a generic extractor into a scenario-expert without human intervention.

To address these challenges, we propose \textbf{Hyper-KGGen}, a novel skill-driven framework designed for high-quality hypergraph generation. 
First, to ensure structural integrity, we introduce a \textbf{Coarse-to-Fine Extraction} mechanism. This module initially establishes a skeletal graph of entities and binary relations, then progressively enriches it with spatiotemporal details to form complex $n$-ary hyperedges, realizing full-dimensional knowledge modeling. 
Second, to overcome the scenario gap, we propose an \textbf{Adaptive Skill Acquisition} mechanism. Instead of relying on static prompts, this module actively distills high-quality extraction skills from the execution history of model, evolving a Global Skill Library $\mathcal{S}$. 
Finally, tackling the difficulty of quantifying scenario-specific knowledge, we design a \textbf{Stability-based Relative Reward} strategy. By categorizing extraction results into stable, unstable, and missed sets, we quantify the model's confidence boundaries, enabling targeted optimization through path induction and hindsight reasoning.
To facilitate research in this area, we also construct and release \textbf{HyperDocRED}, a high-quality benchmark for document-level hypergraph extraction. Extensive experiments demonstrate that Hyper-KGGen significantly outperforms state-of-the-art baselines, verifying that learned skills carry substantially more informational value than few-shot examples in multi-scenario settings.

Our contributions are summarized as follows:
\begin{itemize}[leftmargin=*, noitemsep, topsep=2pt]
    \item We propose a Coarse-to-Fine extraction paradigm that systematically decomposes complex documents, achieving robust full-dimensional hypergraph modeling from binary skeletons to $n$-ary details.
    \item We introduce a Skill-Driven mechanism to bridge the scenario gap. By maintaining a Global Skill Library, our method enables the model to actively acquire and deploy domain-specific expertise, overcoming the brittleness of generic prompts.
    \item We design a Stability-based Relative Reward strategy to quantify the elusive concept of scenario knowledge. By leveraging unstable and missed extraction sets, we implement an efficient self-improving loop that minimizes domain blind spots.
    \item We release HyperDocRED, a rigorously annotated benchmark for n-ary knowledge extraction, to foster further research in high-quality knowledge hypergraph construction.
\end{itemize}

%% file: section/2_rel.tex
\section{Related work}
\subsection{Knowledge Extraction}
Knowledge Extraction is a fundamental task in Natural Language Processing, typically represented through the canonical structure of triple-based Knowledge Graphs. Early works~\cite{suchanek2007yago,oramas2015rule, norabid2022rule} primarily relied on hard-coded heuristic rules to extract knowledge from multi-domain corpora. With the advancement of deep learning, language models based on the Transformer architecture~\cite{arsenyan2024large,zhang2024extract, qiao2022joint} have gradually superseded traditional encoding approaches. This paradigm evolution catalyzed the development of automated Knowledge Graph construction frameworks, which are capable of capturing intricate semantic relationships between entities. More recently, the emergence of Large Language Models (LLMs) has facilitated efficient zero-shot extraction. Some methods~\cite{mo2025kggen,zhang2025rakg, huang2025can} have explored orchestrating LLM-based workflows for knowledge extraction, which boost both efficiency and generalization across diverse texts.
However, triple-based methods are limited to binary relations, making it difficult to cover complex, multi-entity facts and their real-world semantic details.

\subsection{Hypergraph Knowledge Extraction}
To better capture $n$-ary facts and qualifiers beyond triple-based graphs, recent work has increasingly adopted knowledge hypergraphs as a more faithful representation for knowledge extraction. Early works~\cite{chia2022dataset}, proposing the concept of hyper-relation, model argument and qualifier interactions via a cube-filling formulation, while Text2NKG~\cite{luo2024text2nkg} further casts $n$-ary extraction as multi-label classification over span tuples and supports multiple schemas. However, these approaches often rely on predefined label sets, limiting coverage on diverse genres and long-tail relations.

With the improved generalization of LLMs, recent RAG systems have started to construct knowledge hypergraphs directly from raw text using LLM-driven pipelines. Hyper-RAG~\cite{feng2025hyper} adopts a two-stage process: it first extracts entities and then induces both low-order (binary) and high-order (multi-entity) associations as hyperedges. HyperGraphRAG~\cite{luo2025hypergraphrag} instead treats each text chunk as a hyperedge-like knowledge unit and links mentioned entities as incident nodes, while Cog-RAG~\cite{hu2025cog} further introduces a theme-level hypergraph with thematic summarization to enhance semantic coverage, enabling efficient and high-quality hypergraph extraction across diverse textual scenarios.
Following this direction in hypergraph-based knowledge modeling, HGMEM represents multi-step RAG memory as a dynamic hypergraph, where evolving hyperedges act as knowledge units and capture high-order associations among retrieved facts~\cite{zhou2025improving}.

Despite their strong performance on complex texts, existing methods typically rely on fixed extraction strategies and often fail to produce consistently high-quality hypergraphs when transferred across domains due to scenario gaps. Hyper-KGGen introduces an \textit{adaptive skill acquisition} module that actively distills domain expertise into a Global Skill Library and applies these skills during extraction to overcome this limitation.
\subsection{Skill Distillation and Reuse}
Skill distillation and reuse treat the reasoning traces and execution processes produced at test time as a learnable source of skills, and distill these skills into reusable external priors. This line of research is closely related to an agent’s reflection mechanism and related forms of memory-augmented, test-time adaptation.

Reflexion ~\cite{shinn2023reflexion} stores self-reflections as episodic memory and reuses them to steer later attempts. DSPy~\cite{khattab2023dspy} compiles prompts and intermediate modules into a searchable, compositional program and selects configurations that yield more stable reasoning. Recent work further distills test-time rollouts into reusable repositories. ACE~\cite{zhang2025agentic} curates a strategy library via generate–reflect–filter cycles and retrieves strategies for new queries. Training-Free GRPO~\cite{cai2025training} optimizes context-level experience entries through explicit edits rather than parameter updates. RSE~\cite{wang2026not} aggregates test-time samples into an experience bank by extracting stable intermediate conclusions and recurring failure patterns.

Although these methods have demonstrated the ability to improve continuously in general reasoning and agentic workflow settings, whether their underlying ideas can be effectively transferred to knowledge extraction remains unexplored. Hyper-KGGen fills this gap by learning transferable reasoning skills across scenarios from historical extraction traces, and summarizing LLM-intrinsic logical blind spots to drive more reliable hypergraph generation.

%% file: section/3_pre.tex
\section{Preliminary and Definition}
\label{sec:preliminary}
In this section, we formally define the \textbf{Knowledge Hypergraph Extraction Task} and specify the structural representation of entities and relations. Furthermore, we establish the \textbf{Optimization Objective}, aiming to align the generated hypergraph with the human-annotated gold standard, thereby setting the stage for our proposed framework.

\vpara{Knowledge Hypergraph Extraction Task.}
Given a raw document $D$, the goal is to reconstruct the underlying semantic structure by mapping the unstructured text into a structured knowledge hypergraph:
\begin{equation}
    \mathcal{G} = (\mathcal{V}, \mathcal{E}),
\end{equation}
where the node set $\mathcal{V}$ corresponds to the unique entities extracted from Document $D$, and the hyperedge set $\mathcal{E}$ represents the atomic knowledge units connecting these entities.
Unlike conventional knowledge graphs where edges are strictly binary, a hyperedge $e \in \mathcal{E}$ in our setting is defined as a tuple to capture complex high-order $n$-ary interactions:
\begin{equation}
    e=(r, \mathcal{V}_e),
\end{equation}
where $r$ denotes the semantic description (or relation type) of the knowledge unit, and $\mathcal{V}_e \subseteq \mathcal{V}$ denotes the subset of entities involved in this relation. To unify both simple binary relations and complex multi-party interactions within a single framework, we constrain the cardinality of the entity subset to satisfy $|\mathcal{V}_e| \ge 2$.

\vpara{Optimization Objective.}
We assume the existence of an ideal, ground-truth knowledge hypergraph $\mathcal{G}^*=(\mathcal{V}^*, \mathcal{E}^*)$ for a given document $D$. This gold standard is characterized by being \textit{comprehensive} (covering all salient facts in $D$) and \textit{precise} (accurately reflecting semantic dependencies).
Consequently, the extraction task is formalized as learning a mapping function $f$:
\begin{equation}
    \hat{\mathcal{G}} = f(D; \theta),
\end{equation}
where $\theta$ represents the parameters of the extraction model. The ultimate objective is to generate a predicted hypergraph $\hat{\mathcal{G}} = (\hat{\mathcal{V}}, \hat{\mathcal{E}})$ such that it maximizes the alignment with the ideal graph $\mathcal{G}^*$. Specifically, this requires maximizing the semantic overlap between the predicted hyperedges $\hat{\mathcal{E}}$ and the ground-truth hyperedges $\mathcal{E}^*$, ensuring that complex $n$-ary relationships are correctly identified and structurally formed.

%% file: section/4_method.tex
\begin{figure*}[t]
    \centering
    \includegraphics[width=1\linewidth]{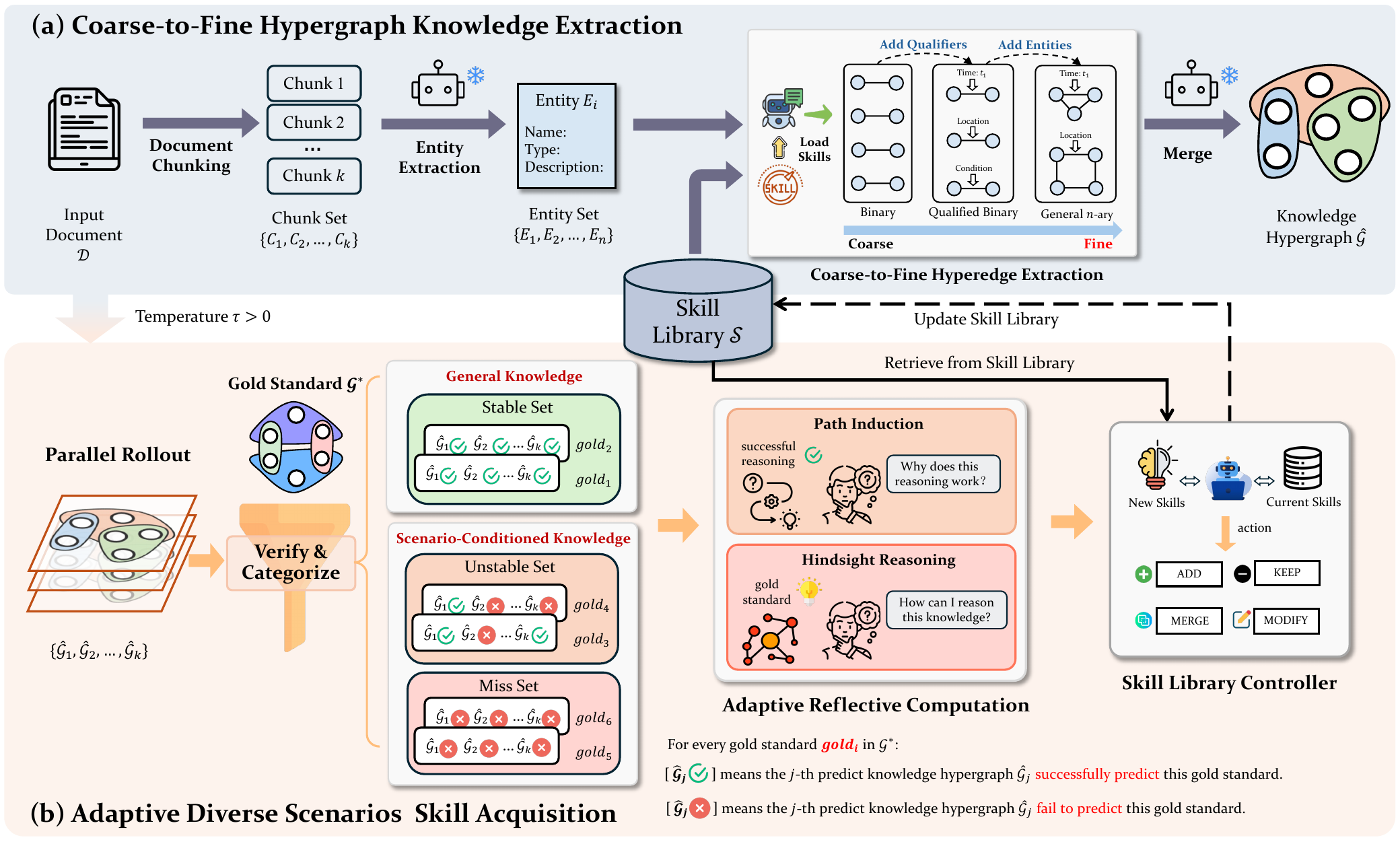}
    \caption{The overall architecture of our proposed Hyper-KGGen framework for high quality knowledge hypergraph generation. It consists two modules: (a) is the Coarse-to-Fine Knowledge Hypergraph Extraction module, and (b) is the Adaptive Diverse Scenarios Skill Acquisition module for iteratively generate reusable skills to Skill Library from execution history.}
    \label{fig:main}
\end{figure*}

\section{Methodology}
\label{sec:methodology}

In this section, we present \textbf{Hyper-KGGen}, a novel framework designed for high-quality knowledge hypergraph construction across diverse scenarios. Unlike prior approaches that rely on rigid schema matching or extensive parameter fine-tuning, Hyper-KGGen integrates a hierarchical structural parsing mechanism with a dynamic skill evolution strategy. This allows the model to systematically decompose complex documents while actively acquiring scenario-specific extraction skills from its own successes and failures.

\subsection{Overview}
\label{sec:overview}
To achieve the optimization objective defined in \cref{sec:preliminary}, we reformulate the knowledge extraction task as a conditional generation process augmented by an external evolving memory. Instead of updating the massive model parameters $\theta$ via computationally expensive gradient descent, we freeze the LLM and optimize a semantic \textbf{Global Skill Library} $\mathcal{S}$.
Formally, the generation process is defined as:
\begin{equation}
    \hat{\mathcal{G}} = f(D, \mathcal{S}; \theta),
\end{equation}
where $D$ denotes the input document, $\theta$ represents the frozen LLM parameters, and $\mathcal{S}$ is learnable skill library, initialized as $\mathcal{S}_0 = \emptyset$.

Our goal is to iteratively evolve $\mathcal{S}$ such that the generated hypergraph $\hat{\mathcal{G}}$ progressively aligns with the gold standard $\mathcal{G}^*$. As illustrated in \cref{fig:main}, the framework operates in a closed-loop cycle comprising two core phases:

\begin{itemize}[leftmargin=*, noitemsep, topsep=2pt]
    \item \textbf{Forward Pass: Coarse-to-Fine Extraction.} 
    Given a document $D$ and the current skills $\mathcal{S}$, the model first establishes a structural skeleton of entities and binary relations, then progressively fills in spatiotemporal details to construct complex $n$-ary hyperedges. This hierarchical approach reduces the cognitive load on the LLM, ensuring the structural integrity of the output graph $\hat{\mathcal{G}} = (\mathcal{V}, \mathcal{E})$.
    
    \item \textbf{Backward Feedback: Adaptive Skill Acquisition.} 
    To bridge the gap between the generated $\hat{\mathcal{G}}$ and the ground truth $\mathcal{G}^*$, we design a feedback mechanism based on extraction stability. By performing parallel rollouts, we categorize relations into stable, unstable, and missed sets. The system then distills high-quality extraction skills from these discrepancies—specifically through path induction for unstable relations and hindsight reasoning for missed ones—and updates the Global Skill Library $\mathcal{S}$ via a memory controller.
\end{itemize}

Through this iterative interplay, Hyper-KGGen continuously accumulates scenario-specific expertise, enabling robust and high-precision knowledge extraction across varying domains.

\subsection{Coarse-to-Fine Hypergraph Knowledge Extraction}
\label{sec:extraction}
In this module, we propose a Coarse-to-Fine Knowledge Extraction pipeline designed to transform unstructured documents into a high-quality knowledge hypergraph $\mathcal{G} = (\mathcal{V}, \mathcal{E})$. Unlike traditional approaches that treat all relations uniformly, our method follows a hierarchical paradigm: it first establishes the structural skeleton of the text and progressively fills in spatiotemporal details and complex narrative events. The overall process comprises four distinct stages: document chunking, entity extraction, coarse-to-fine hyperedge extraction, and global knowledge deduplication.

\subsubsection{Document Chunking}
\label{sec:chunking}
Directly processing long documents poses challenges for LLMs due to context window limitations. To address this, we employ an adaptive document chunking strategy that preserves semantic integrity. Instead of applying fixed-length truncation, we split the input document $D$ at natural boundaries (e.g., sentence endings or paragraph breaks).
Formally, we generate a sequence of chunks:
\begin{equation}
    \mathcal{C} = \{c_1, c_2, \dots, c_m\}.
\end{equation}
To mitigate information loss at cut-off points, we introduce an overlapping window mechanism where adjacent chunks share a localized context $o$. This redundancy ensures that cross-boundary entities and relations are captured in at least one chunk, facilitating subsequent extraction tasks.

\subsubsection{Entity Extraction}
\label{sec:entity_extract}
Entities serve as the fundamental nodes $\mathcal{V}$ in the hypergraph $\mathcal{G}$. For each chunk $c_i \in \mathcal{C}$, we prompt the LLM to identify all valid mentions, assign fine-grained types, and generate concise descriptions based on the local context. This step isolates the atomic semantic units of the document, creating a candidate entity pool necessary for constructing complex relationships.

\subsubsection{Coarse-to-Fine Hyperedge Extraction}
\label{sec:hyperedge_extract}
Drawing upon the extracted entities, we design a \textit{Coarse-to-Fine} modeling strategy to mine hyperedges $\mathcal{E}$. We categorize relationships into three granularities, guiding the model to progressively capture information from the basic structural skeleton to intricate narrative details:

\begin{itemize}[leftmargin=*, noitemsep, topsep=2pt]
    \item \textbf{Binary Relations (The Skeleton).} These represent the coarsest level of interaction, capturing the fundamental pairwise links between entities (e.g., \textit{Subject-Predicate-Object}). They serve as the ``structural skeleton'' of the hypergraph, providing the basic connectivity without complex constraints.
    
    \item \textbf{Qualified Binary Relations (Contextual Augmentation).} Moving to a finer granularity, these relations augment standard binary links with qualifying arguments such as time, location, or specific conditions. This category transforms static binary links into dynamic interactions situated in specific spatiotemporal contexts, offering a richer representation than simple triples.
    
    \item \textbf{General $N$-ary Relations (Event Details).} At the finest level, we model complex interactions involving multiple entities as general hyperedges in $\mathcal{E}$. Unlike binary links, these relations encapsulate entire events or story plots, where multiple participants jointly instantiate a coherent scenario. This granularity captures the nuances of the narrative, treating events as holistic structured units rather than decomposing them into fragmented pairwise edges.
\end{itemize}

\subsubsection{Knowledge Deduplication}
\label{sec:deduplication}
Since the chunking process involves overlaps and entities may appear repeatedly across different chunks, the raw extraction results contain redundancy. We implement a global clustering and fusion mechanism to consolidate the knowledge.
First, we perform cross-chunk coreference resolution to cluster mentions referring to the same real-world entity. Second, we merge hyperedges that share identical semantic meanings. For both entities and relations, we synthesize their descriptions by aggregating details from all instances, resulting in a unified and low-redundancy knowledge hypergraph $\mathcal{G}$.

\subsection{Adaptive Skill Acquisition for Diverse Scenarios}
\label{sec:adaptive_skill}
While LLMs possess extensive general knowledge, they often struggle to align with the intricate constraints of specific domains, particularly when extracting complex $n$-ary hyperedges. To bridge this gap, we propose an Adaptive Skill Acquisition mechanism. Instead of static prompt tuning, this module establishes a dynamic feedback loop that categorizes extraction results based on their stability, efficiently distilling high-quality extraction skills from unstable and missed instances to refine a global skill library $\mathcal{S}$.

\subsubsection{Parallel Rollout for Candidate Generation}
\label{sec:parallel_rollout}
Extracting complex $n$-ary relations involves high uncertainty. Single-pass inference is susceptible to decoding randomness and cognitive bias, often failing to cover the intricate semantic structures hidden in the document. To mitigate this, we employ a Parallel Rollout strategy.
Given a training document $D$ and the current skill library $\mathcal{S}$, we independently sample the model's output $K$ times using a non-zero temperature $T$. This generates a diverse set of candidate hypergraphs to probe the model's capability boundaries:
\begin{equation}
\hat{\mathcal{G}}^{(k)} \sim p_{\theta}\!\left(\mathcal{G} \mid D, \mathcal{S}; T\right),\quad k=1,\ldots,K,
\end{equation}
\begin{equation}
    \hat{\mathcal{G}}(D)=\left\{\hat{\mathcal{G}}^{(1)},\ldots,\hat{\mathcal{G}}^{(K)}\right\}.
\end{equation}
Here, $\hat{\mathcal{G}}(D)$ aggregates $K$ parallel hypotheses. This multi-path exploration helps offset random perturbations and exposes the stability of different relational patterns, providing a rich data source for subsequent analysis.

\subsubsection{Adaptive Reward for Diverse Scenarios}
\label{sec:adaptive_reward}
Evaluating $n$-ary relations is significantly more challenging than binary relations due to their fuzzy boundaries and multi-argument complexity. Hard matching often fails to capture partial successes (e.g., extracting 9 out of 10 nodes). Therefore, we introduce a soft evaluation metric based on semantic distance. By computing the embedding similarity between the descriptions of extracted hyperedges and the gold standard $\mathcal{G}^*$, we assess extraction quality in a continuous semantic space.
Furthermore, to quantify the distinction between general knowledge and scenario-specific knowledge, we implement a stability-based reward strategy.

\vpara{Stability-based Relative Reward.}
By aligning the $K$ candidate graphs with $\mathcal{G}^*$, we categorize the gold relations into three distinct subsets based on their retrieval frequency:
\begin{itemize}[leftmargin=*, noitemsep, topsep=2pt]
    \item \textbf{Stable Set.} Relations consistently retrieved across all samples. These correspond to \textit{General Knowledge} that the model's internal weights can handle robustly without external aid.
    \item \textbf{Unstable Set.} Relations retrieved only in a subset of trajectories. These typically represent \textit{Scenario-Specific Knowledge} (e.g., industry jargon or implicit connections), where the model oscillates between success and failure due to insufficient grounding and low confidence.
    \item \textbf{Miss Set.} Relations never retrieved in any sample. These indicate \textit{Domain-Exclusive Knowledge} or deep reasoning gaps where the model completely lacks the necessary extraction logic.
\end{itemize}

\vpara{Adaptive Reflective Computation.}
Based on the categorization above, we ignore the Stable Set (as no optimization is needed) and focus on distilling skills from the latter two:
\begin{itemize}[leftmargin=*, noitemsep, topsep=2pt]
    \item \textbf{Path Induction for Unstable Set.} For unstable relations, we analyze the successful trajectories where the model correctly extracted the hyperedge. We prompt the LLM to summarize the reasoning path that led to these successes, explicitly articulating the scenario-specific logic to stabilize future inference.
    \item \textbf{Hindsight Reasoning for Miss Set.} For relations in the Miss Set, we employ a \textit{Hindsight Reasoning} strategy. We inject the ground-truth relation from $\mathcal{G}^*$ into the context as a posterior condition. Knowing that the relation exists, the model backtracks through the document, locates overlooked evidence, and constructs a logical chain from scratch. This process generates a fresh extraction rule aimed at covering the prior blind spots.
\end{itemize}
These reflective processes output a set of optimization proposals, which are then fed into the library maintenance module.

\subsubsection{Adaptive Skill Acquisition and Deployment}
\label{sec:skill_deploy}
This module consolidates the fragmented proposals into the global skill library $\mathcal{S}$ and deploys them to enhance multi-scenario extraction.

\vpara{Skill Acquisition.}
To maintain the efficiency of $\mathcal{S}$, we employ a Skill Library Controller that ensures compactness and consistency. It processes incoming proposals via four dynamic operations: 
\begin{itemize}
[leftmargin=*,itemsep=0pt,parsep=0.2em,topsep=0.3em,partopsep=0.3em]
\item \textbf{ADD} new skills to cover missed extraction patterns;
\item \textbf{MODIFY} existing skills to improve generalizability;
\item \textbf{MERGE} redundant skills to maintain a compact skill library;
\item \textbf{KEEP} the skill unchanged.
\end{itemize}
This ensures that $\mathcal{S}$ evolves into a high-quality repository of distinct extraction skills.

\vpara{Skill Deployment.}
During the inference phase, relevant skills from $\mathcal{S}$ are dynamically retrieved and injected into the prompt. This augments the generic instructions with scenario-specific expertise, enabling the model to perform high-quality hypergraph generation across diverse domains.

%% file: section/5_exp.tex
\section{Experiment}
In this section, we present the experimental evaluation of Hyper-KGGen.
We report result for two variants: {Hyper-KGGen} denotes the base model without acquired skills while {Hyper-KGGen$^{+}$} represents the skill-augmented model that incorporates the learned skill library $\mathcal{S}$.
We first outline the experimental setup and compare Hyper-KGGen with various baselines, then provide an in-depth analysis of its hypergraph generation ability, focusing on $n$-ary relation extraction, factual coverage, and downstream utility.

\begin{figure*}[t]
    \centering
    \includegraphics[width=0.95\linewidth]{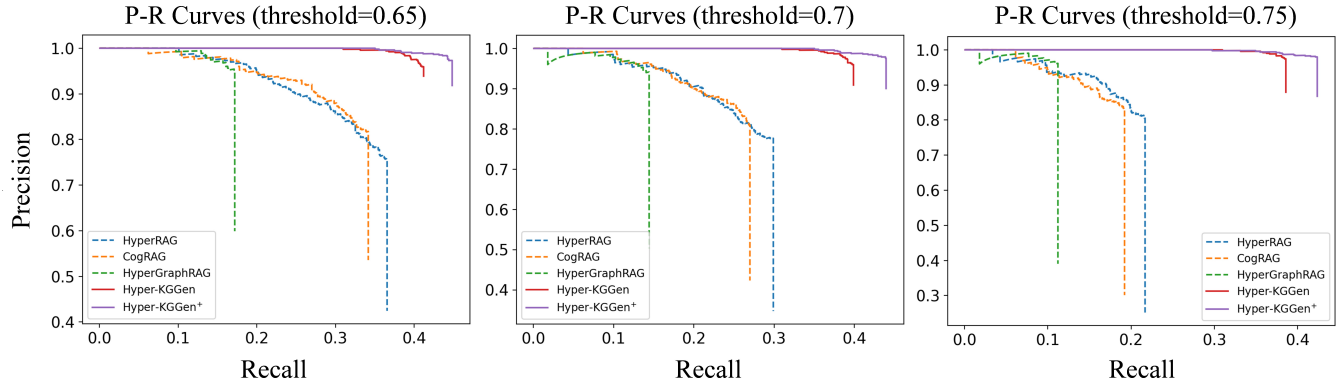}
    \caption{Precision-Recall Curves for $n$-ary Relation Extraction on the HyperDocRED Dataset.}
    \label{fig:pr}
\end{figure*}

\subsection{Experiment Setting}
\subsubsection{Dataset}
To evaluate the ability of Hyper-KGGen in handling diverse document structures and complex semantic relations, we conduct experiments on three tasks across four datasets, including a newly constructed benchmark for $n$-ary relation extraction:

\begin{table}[t]
\centering
\small
\setlength{\tabcolsep}{6pt}
\renewcommand{\arraystretch}{1.1}
\caption{Statistics of the HyperDocRED Dataset.}
\label{tab:dataset_stats}
\resizebox{0.46\textwidth}{!}{%
\begin{tabular}{lccccc}
\toprule
\multirow{2}{*}{\textbf{Split}} &
\multirow{2}{*}{\textbf{Sample}} &
\multirow{2}{*}{\textbf{Entity}} &
\multicolumn{3}{c}{\textbf{Correlation}} \\
\cmidrule(lr){4-6}
& & & \textbf{Low-order} & \textbf{High-order} & \textbf{Total} \\
\midrule
Train & 50  & 1016 & 326 & 310 & 636  \\
Test  & 100 & 2127 & 758 & 613 & 1371 \\
\bottomrule
\end{tabular}%
}
\end{table}

\vpara{HyperDocRED Dataset:} To evaluate the performance on a standard benchmark adapted for $n$-ary relational contexts, we construct the HyperDocRED Bench based on the widely used Re-DocRED corpus \cite{tan2022revisiting}. We manually restructure the original binary annotations into $n$-ary relations to match the requirements of knowledge hypergraph construction. Specifically, we target explicit and complete $n$-ary relations as expressed in the text, preserving each relation as an integral semantic unit rather than fragmenting it into multiple binary triples. The dataset contains 50 training documents serving as the seed source for skill acquisition and 100 test documents for evaluation. Detailed statistics are shown in \cref{tab:dataset_stats}.

\vpara{MINE Dataset} \cite{mo2025kggen}: The MINE dataset serves as a benchmark for complex document-level extraction, containing 100 articles. The corpus is linguistically diverse, covering multiple domains including history, art, science, ethics, and psychology. The articles are generated by the LLM based on 100 distinct topics to ensure semantic richness, followed by rigorous human verification to establish the gold standard for hyper-relational extraction.

\vpara{UltraDomain Benchmark} \cite{qian2025memorag}: The UltraDomain is sourced from 428 college textbooks and encompasses 18 distinct domains, covering fields such as social sciences and the humanities. We choose the Pathology and Mix datasets for the downstream RAG evaluation.

\subsubsection{Baselines}
In performance comparison, we consider several state-of-the-art baselines, including knowledge graph extraction models: KGGen \cite{mo2025kggen}, iText2KG \cite{lairgi2024itext2kg}, RAKG \cite{zhang2025rakg}, text-based RAG: NativeRAG ~\cite{gao2023retrieval}, graph-based RAG: GraphRAG ~\cite{edge2024local}, LightRAG ~\cite{guo2024lightrag}, HiRAG ~\cite{huang2025retrieval}, and hypergraph-enhanced RAG approaches: HyperGraphRAG \cite{luo2025hypergraphrag}, Hyper-RAG \cite{feng2025hyper}, and Cog-RAG \cite{hu2025cog}. We follow the official settings and implementations with GPT-4o-mini as the LLM for all baselines. Baseline details are provided in Appendix \ref{app:baseline}.

\begin{table*}[t]
\centering
\small
\setlength{\tabcolsep}{6pt}
\renewcommand{\arraystretch}{1.1}
\sisetup{
  detect-weight=true,
  detect-inline-weight=math,
  table-number-alignment=center
}
\caption{Instructional Response Quality on Mix and Pathology across Five Dimensions.}
\label{tab:mix_pathology_quality}
\begin{tabular}{l
                |S[table-format=2.2] S[table-format=2.2] S[table-format=2.2] S[table-format=2.2] S[table-format=2.2] S[table-format=2.2]|
                S[table-format=2.2] S[table-format=2.2] S[table-format=2.2] S[table-format=2.2] S[table-format=2.2] S[table-format=2.2]}
\toprule
\multirow{2}{*}{\textbf{Method}} &
\multicolumn{6}{c}{\textbf{Mix}} &
\multicolumn{6}{|c}{\textbf{Pathology}} \\
\cmidrule(lr){2-7}\cmidrule(lr){8-13}
& {\textbf{Comp.}} & {\textbf{Diver.}} & {\textbf{Empo.}} & {\textbf{Logi.}} & {\textbf{Read.}} & {\textbf{Avg.}}
& {\textbf{Comp.}} & {\textbf{Diver.}} & {\textbf{Empo.}} & {\textbf{Logi.}} & {\textbf{Read.}} & {\textbf{Avg.}} \\
\midrule
NativeRAG      & 87.10 & 76.64 & 74.62 & 82.06 & 80.46 & 80.18 & 90.50 & 83.60 & 77.00 & 86.50 & 86.50 & 84.22 \\
GraphRAG       & 89.00 & 79.70 & 74.66 & 83.00 & 82.32 & 81.74 & 90.60 & 82.20 & 78.78 & 87.20 & 86.42 & 85.04 \\
LightRAG       & 88.20 & 75.90 & 74.94 & 83.80 & 82.92 & 81.15 & 91.50 & 83.74 & 78.70 & 86.76 & 85.58 & 85.26 \\
HiRAG          & 88.00 & 79.86 & 76.80 & 84.06 & 83.44 & 82.43 & 90.90 & 83.56 & \underline{80.98} & 87.16 & 86.34 & 85.79 \\
Hyper-RAG      & \underline{90.50} & \textbf{83.40} & 78.32 & 85.40 & 83.96 & 84.32 & 91.40 & 84.92 & 80.60 & 87.22 & 86.00 & 86.03 \\
HyperGraphRAG  & 90.20 & 81.50 & 78.70 & 84.94 & 83.46 & 83.76 & 91.70 & 83.64 & 79.96 & 87.54 & 86.34 & 85.84 \\
Cog-RAG        & 90.10 & \underline{83.34} & 78.52 & \underline{85.94} & 84.96 & 84.57 & 91.10 & \underline{85.30} & 79.02 & \underline{88.30} & \underline{86.84} & 86.11 \\
\midrule
\textbf{Hyper-KGGen} & 90.02 & 82.10 & \underline{79.04} & \textbf{86.78} &  \underline{85.42} & \underline{84.67} & \underline{91.90} & 85.00 & 79.70 & \textbf{88.50} & \textbf{87.38} & \underline{86.50} \\
\textbf{Hyper-KGGen$^{+}$} &
\textbf{91.24} & 83.00 & \textbf{79.91} & 85.61 & \textbf{85.74} & \textbf{85.10} & \textbf{92.35} & \textbf{85.71} & \textbf{81.10} & 88.27 & 86.17 & \textbf{86.72} \\
\bottomrule
\end{tabular}%
\end{table*}

\subsubsection{Implementation Details}
We evaluate Hyper-KGGen along three complementary axes, covering $n$-ary relation extraction, fact coverage, and downstream utility.
(1) For $n$-ary relation extraction on HyperDocRED, the precision and recall of $n$-ary relations are measured using a semantic-based evaluation pipeline: relation descriptions are encoded with the all-MiniLM-L6-v2 SentenceTransformer. We then compute a cosine-similarity matrix and apply the Hungarian algorithm to find the optimal global matching between predicted and ground-truth relations. 
(2) For fact coverage on MINE, we formulate a fact-verification task with 15 ground-truth facts per article. For each query fact, we retrieve the top 5 nodes with the highest semantic similarity and expand each retrieved node to its 2-hop neighborhood to construct a candidate subgraph. The retrieved contexts are provided to an LLM, which performs binary fact-checking conditioned on the assembled context evidence (1 if the retrieved context supports the fact, 0 otherwise).
(3) For downstream utility on UltraDomain, we integrate the Hyper-KGGen as the knowledge hypergraph database construction backbone into the standard RAG pipeline on downstream benchmarks. Specifically, we construct a structured hypergraph database from the corpus with a standard procedure.
At retrieval time, given a user query, we retrieve the relevant entities and hyperedges, perform a one-hop expansion to collect context evidence, and then feed the retrieved evidence to the LLM for response generation.

\subsubsection{Evaluation Metrics}
For $n$-ary relation extraction, we plot \textbf{Precision-Recall} curves by treating each predicted $n$-ary relation as a detection and each gold relation as a ground-truth instance.
For fact coverage, we verify ground-truth facts and report \textbf{Accuracy} over all facts in the test set.
For downstream utility, we adopt an LLM-based \textbf{score evaluation} strategy \cite{wang2024leave,feng2025hyper} that judges the responses produced by baselines along five dimensions: Comprehensiveness, Diversity, Empowerment, Logical, and Readability.

\subsection{Main Results}
In this section, we present a comprehensive evaluation of Hyper-KGGen compared with state-of-the-art baselines.
To assess its capabilities from complementary perspectives, we organize our experiments into three aspects: 
(1) \textbf{$n$-ary relation extraction}, evaluating the structural accuracy of the generated $n$-ary relations on the HyperDocRED dataset; (2) \textbf{Fact coverage}, measuring semantic completeness and information retention on MINE dataset; and (3) \textbf{Downstream utility}, validating whether the constructed hypergraphs improve RAG task performance on UltraDomain benchmarks.

\vpara{Superior Performance in $n$-ary Relation Extraction.}
As shown in \cref{fig:pr}, Hyper-KGGen consistently outperforms all baseline methods.
Hyper-KGGen achieves exceptional recall for $n$-ary relations while maintaining high precision, indicating its ability to capture complete $n$-ary semantics rather than fragment them into arbitrary components. Furthermore, the deployment of learned skills further uncovers additional $n$-ary relations without compromising precision, which validates the effectiveness of skills distilled from different scenario knowledge during adaptive skill acquisition.

\begin{figure}[t]
    \centering
    \includegraphics[width=\linewidth]{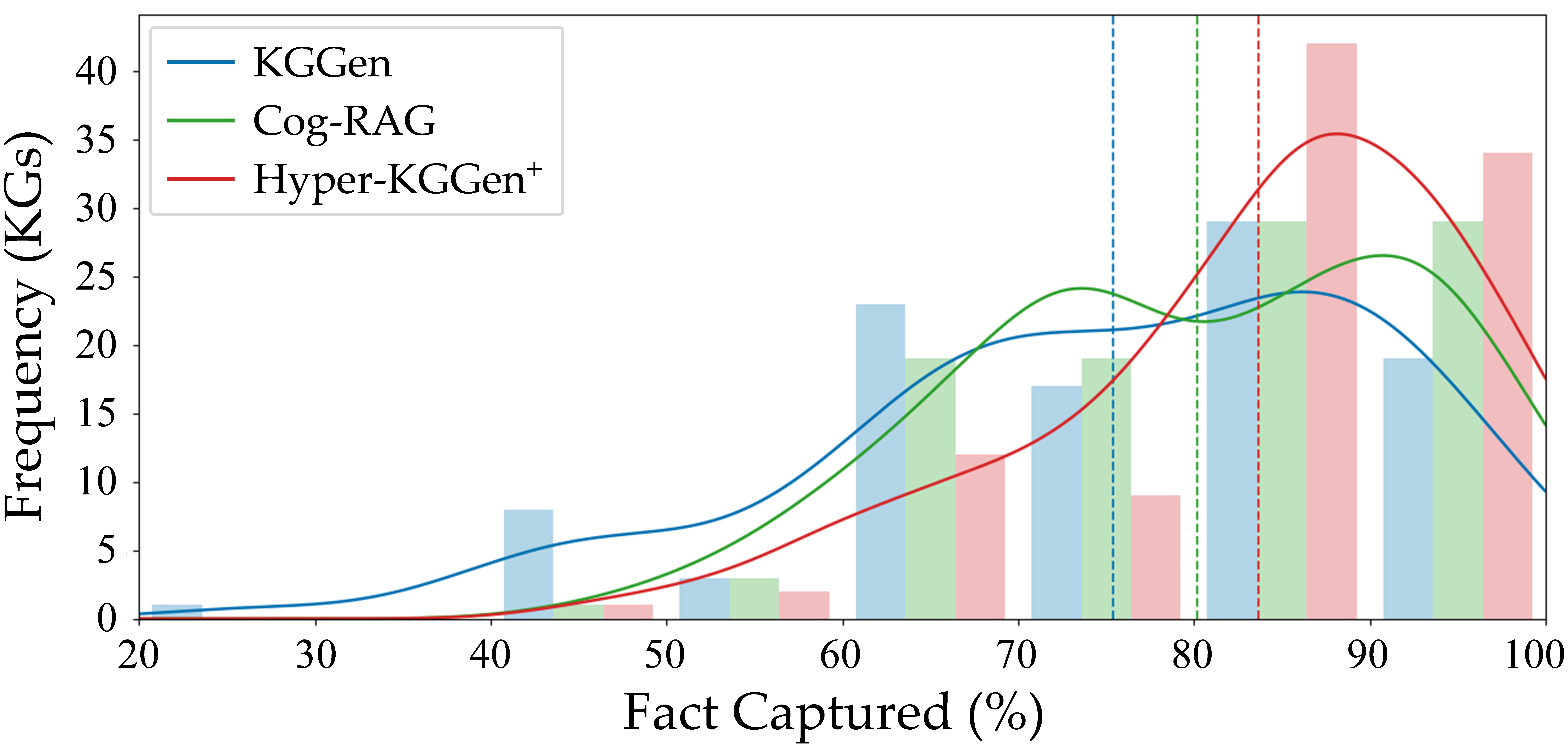}
    \caption{Distribution of MINE scores across 100 articles for KGGen, Cog-RAG, and Hyper-KGGen. Dotted vertical lines show average performance.}
    \label{fig:mine}
\end{figure}

\vpara{High Fact Coverage Across Domains.} 
Beyond its strong performance in $n$-ary relation extraction, Hyper-KGGen captures a broader set of salient facts from diverse-domain texts.
As shown in \cref{fig:mine}, the fact-capture distribution of Hyper-KGGen is clearly shifted toward higher coverage: a larger fraction of its constructed KGs concentrate in the high-capture regime, while the baselines exhibit a heavier low-coverage tail and more mass in mid-range bins. 
The dotted vertical line of Hyper-KGGen also lies rightmost, indicating a higher overall coverage. 
These results suggest that the acquired skills provide additional scenario priors, improving robustness across domains and enabling a more effective structured representation of the source text with reduced information loss.

\vpara{Downstream Utility on RAG Tasks.}
The effectiveness of structured RAG systems depends not only on retrieval heuristics, but also on the quality of the underlying knowledge substrate. To further examine how the generated structured knowledge contributes to downstream utility, we conduct RAG experiments. As shown in \cref{tab:mix_pathology_quality}, Hyper-KGGen achieves the best average instructional quality on Mix and Pathology.
This reveals the reliability of the knowledge hypergraphs generated by Hyper-KGGen and suggests that improving the hypergraph substrate can boost utility even with simple retrieval, highlighting that high-quality hypergraph construction is a primary driver of downstream utility.
Notably, the gap between {Hyper-KGGen} and baselines is modest, since the graph mainly serves retrieval while the generation relies on raw text chunks.
With sufficient retrieval budget, baselines tend to retrieve highly overlapping text, narrowing the quality gap.
\subsection{Analysis}
In this section, we further analyze the model’s $n$-ary relation extraction performance and its robustness across different backbone LLMs. We also conduct an ablation on the entity retrieval module by varying the top-$k$ retrieved entities and examining the resulting context quality. Finally, we investigate the relationship between skill induction and few-shot prompting in the context of knowledge extraction.

\subsubsection{Semantic Matching Threshold Sensitivity}
We evaluate how varying the semantic matching threshold affects $n$-ary relation extraction performance. As shown in \cref{fig:pr}, we compare precision–recall behavior across models under thresholds of 0.65, 0.70, and 0.75. We observe that Hyper-KGGen consistently maintains high precision regardless of the threshold, indicating its ability to capture complete $n$-ary relations from text without fragmenting them into partial facts. Moreover, across all thresholds, Hyper-KGGen substantially outperforms the baselines, demonstrating robust advantages under different semantic matching criteria.

\begin{table}[t]
\centering
\small
\renewcommand{\arraystretch}{1.1}
\caption{Fact-Verification Accuracy on the MINE Dataset.}
\label{tab:acc_comparison_2}
\setlength{\tabcolsep}{4pt}
\begin{tabular}{lcccc}
\toprule
\textbf{Method} & \textbf{\begin{tabular}[c]{@{}c@{}}GPT-4o \\ -mini\end{tabular}} & \textbf{\begin{tabular}[c]{@{}c@{}}Gemini-2.5 \\ -Flash\end{tabular}} & \textbf{QWen3} & \textbf{DeepSeek-V3.2} \\
\midrule
KGGen          & 0.7540 & 0.6113 & 0.7527 & 0.7273 \\
RAKG           & 0.7659 & 0.6007 & 0.7253 & 0.7486 \\
Hyper-RAG      & 0.8173 & 0.5993 & 0.7873 & 0.7660 \\
HyperGraphRAG & 0.8053 & 0.6780 & 0.7673 & 0.7740 \\
Cog-RAG        & 0.8020 & 0.6013 & 0.7713 & 0.7467 \\
\midrule
\textbf{Hyper-KGGen} & \underline{0.8217} & \underline{0.7060} & \underline{0.8180} & \underline{0.7880} \\
\textbf{Hyper-KGGen$^{+}$}  & \textbf{0.8367} & \textbf{0.7133} & \textbf{0.8473} & \textbf{0.8020} \\
\bottomrule
\end{tabular}%
\end{table}

\subsubsection{Robustness across LLMs}
To examine how the choice of LLM backbone impacts different knowledge extraction methods, we instantiate all models with a diverse set of base LLMs, including GPT-4o-mini \cite{hurst2024gpt}, Gemini-2.5-Flash \cite{comanici2025gemini}, Qwen3 \cite{yang2025qwen3}, and DeepSeek-V3.2 \cite{liu2025deepseek}. As shown in \cref{tab:acc_comparison_2}, Hyper-KGGen consistently extracts more facts across all backbone settings. Moreover, with skill guidance, Hyper-KGGen further improves its robustness across scenarios and enhances extraction efficiency.

\subsubsection{Effect of the Number of Rollouts}
To examine the effect of the number of rollouts, we vary $K$ in adaptive skill acquisition. As shown in \cref{tab:k_hyperdocred_mine}, larger values of $K$ consistently improve recall-oriented metrics and downstream accuracy, suggesting that additional parallel hypotheses provide a more reliable basis for estimating extraction stability. This richer stability signal helps identify unstable and missed high-order relations and supports the distillation of more effective extraction skills. The decrease in precision suggests that broader exploration places greater demands on stability estimation, as more unstable or low-confidence candidates may enter the skill acquisition stage and introduce additional noise into the skill library. Overall, the trend reveals a trade-off between exploration coverage and noise control, with the benefit of additional rollouts gradually saturating as $K$ increases.

\begin{table}[t]
\centering
\small
\renewcommand{\arraystretch}{1.1}
\caption{Performance under Different Rollout Number $K$.}
\label{tab:k_hyperdocred_mine}
\setlength{\tabcolsep}{12pt}
\begin{tabular}{lcccc}
\toprule
\multirow{2}{*}{\textbf{$K$}} & \multicolumn{3}{c}{\textbf{HyperDocRED}} & \textbf{MINE} \\
\cmidrule(lr){2-4} \cmidrule(lr){5-5}
 & \textbf{Precision} & \textbf{Recall} & \textbf{F1} & \textbf{Acc.} \\
\midrule
2 & 0.8232 & 0.4126 & 0.5497 & 0.8314 \\
4 & 0.8142 & 0.4609 & 0.5736 & 0.8367 \\
6 & 0.8116 & 0.4696 & 0.5948 & 0.8402 \\
8 & 0.8078 & 0.4769 & 0.5997 & 0.8422 \\
\bottomrule
\end{tabular}%
\end{table}

\subsubsection{Skills vs. Few-shot}
To understand how external guidance improves knowledge extraction and how it transfers across scenarios, we compare few-shot prompting and skill-driven guidance.

\begin{figure}[t]
    \centering
    \includegraphics[width=\linewidth]{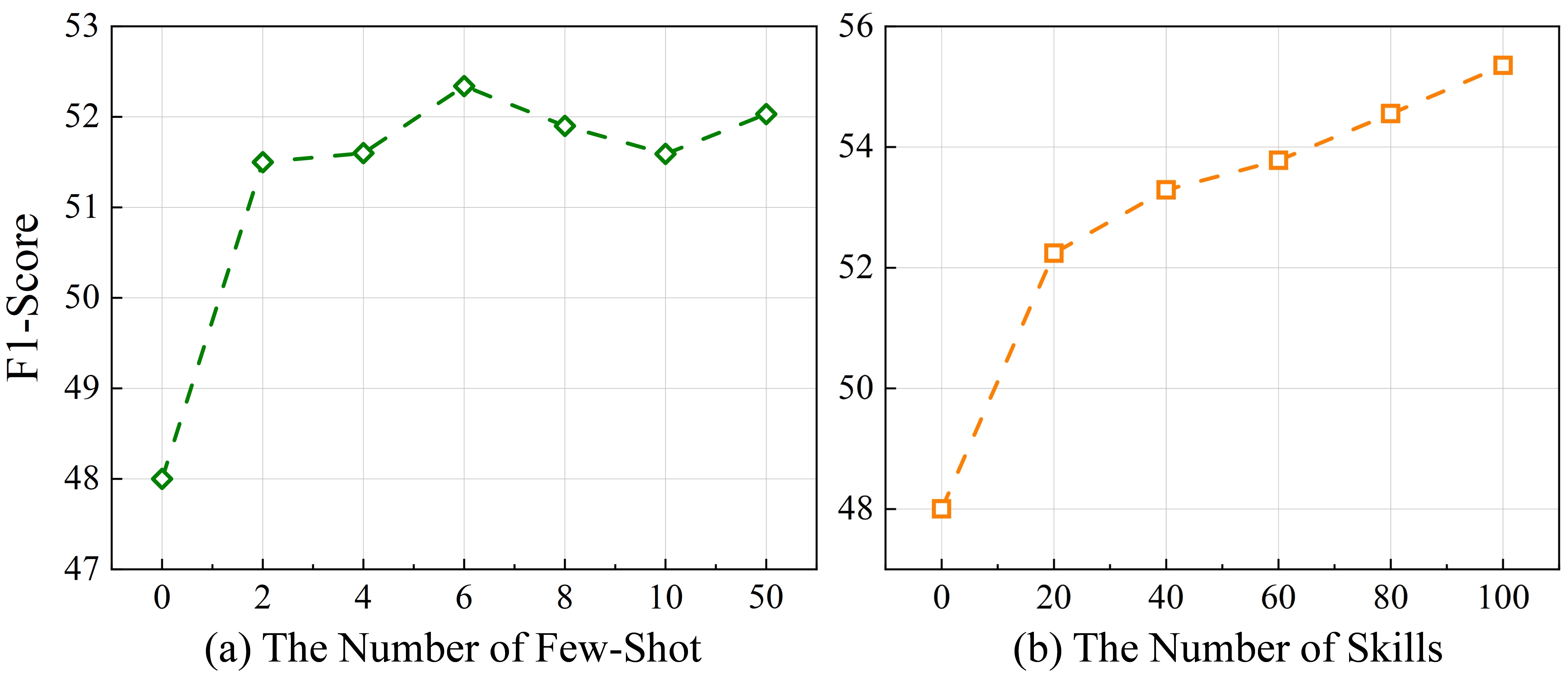}
    \caption{Performance Scaling with Few-Shot Setting and Skill Size on the HyperDocRED Dataset.}
    \label{fig:scaling}
\end{figure}

The results in \cref{fig:scaling} show that both methods outperform the base settings. Few-shot prompting provides a clear boost with a small number of examples. However, adding more examples brings little additional improvement, and performance tends to plateau. This suggests that few-shot prompting mainly helps the model align with the task format and a narrow range of local regularities, but the returns do not keep scaling with more examples.

In contrast, the skill-driven guidance shows steadier and cumulative gains as the library grows, leading to stronger overall performance. Skills are learned from past reasoning traces through a feedback loop that summarizes recurring errors and missing evidence into reusable skills. These skills can then be applied during inference to guide extraction decisions.

A key reason for this gap is that few-shot examples often require full context to be effective, including explicit node connections and surrounding evidence for a relation. Much of this information is irrelevant for a particular query and can weaken the useful signal.
Few-shot prompting is also sensitive to the example distribution. When test cases differ in wording or local subgraph structure, static examples may fail to match the new scenarios. By contrast, skills come from patterns that recur across past cases and scenarios, so they capture decision-relevant constraints and remain effective under scenario shift.

\begin{table}[t]
\centering
\small
\setlength{\tabcolsep}{5pt}
\renewcommand{\arraystretch}{1.12}
\caption{Retrieval Efficiency under Different Top-$k$ Budgets.}
\label{tab:topk_sweep}
\begin{tabular}{l
                S[table-format=2.2]
                S[table-format=2.2]
                S[table-format=2.2]
                S[table-format=2.2]
                S[table-format=2.2]}
\toprule
{\begin{tabular}[l]{@{}l@{}}
\textbf{Factual Coverage} \\ Accuracy \end{tabular}} & {$k=3$} & {$k=5$} & {$k=7$} & {$k=9$} & {$k=11$} \\
\midrule
\multicolumn{6}{l}{\textbf{Return top-$k$ similar entities from the KG in factual coverage}} \\ 
KGGen  & \multicolumn{1}{c}{0.7310} & \multicolumn{1}{c}{0.7540} & \multicolumn{1}{c}{0.7717} & \multicolumn{1}{c}{0.7833} & \multicolumn{1}{c}{0.7887} \\
Hyper-KGGen   & \multicolumn{1}{c}{0.7987} & \multicolumn{1}{c}{0.8217} & \multicolumn{1}{c}{0.8352} & \multicolumn{1}{c}{0.8375} & \multicolumn{1}{c}{0.8417} \\
Hyper-KGGen$^{+}$  & \multicolumn{1}{c}{0.8192} & \multicolumn{1}{c}{0.8367} & \multicolumn{1}{c}{0.8412} & \multicolumn{1}{c}{0.8454} & \multicolumn{1}{c}{0.8487} \\
\midrule
{\begin{tabular}[l]{@{}l@{}}
\textbf{Downstream Utility} \\ Average Score \end{tabular}} & {$k=40$} & {$k=50$} & {$k=60$} & {$k=70$} & {$k=80$} \\
\midrule
\multicolumn{6}{l}{\textbf{Retrieve top-$k$ items from Hypergraph DB within RAG pipeline}} \\ 
Hyper-RAG  & \multicolumn{1}{c}{81.30} & \multicolumn{1}{c}{83.07} & \multicolumn{1}{c}{84.32} & \multicolumn{1}{c}{82.81} & \multicolumn{1}{c}{81.52} \\
Hyper-KGGen & \multicolumn{1}{c}{83.88} & \multicolumn{1}{c}{84.46} & \multicolumn{1}{c}{84.67} & \multicolumn{1}{c}{83.96} & \multicolumn{1}{c}{83.48} \\
Hyper-KGGen$^{+}$  & \multicolumn{1}{c}{83.92} & \multicolumn{1}{c}{84.72} & \multicolumn{1}{c}{85.10} & \multicolumn{1}{c}{84.06} & \multicolumn{1}{c}{83.77} \\
\bottomrule
\end{tabular}
\end{table}

\subsubsection{Retrieval Efficiency Analysis}
Graph quality determines the budget efficiency of graph-based retrieval: under the same retrieval budget, a high-quality graph should cover more key facts with fewer candidate entities and provide denser, more relevant evidence for downstream generation.
Motivated by this, we treat top-$k$ as a retrieval-budget control variable and evaluate the efficiency of evidence retrieval from the generated graph under different $k$ values, thereby indirectly assessing graph construction quality.

As shown in \cref{tab:topk_sweep}, in fact retrieval, Hyper-KGGen consistently achieves a high hit rate across different $k$ settings; even with small $k$, it covers most facts in the source text, indicating higher fact-carrying density and better structural usability of its graph, without relying on large-scale retrieval to compensate for noise or fragmentation.
Furthermore, end-to-end RAG results show that retrieving only a small number of entities (e.g., $k=40$) already yields performance comparable to larger $k$ values and to Hyper-RAG under various configurations. This suggests that Hyper-KGGen delivers higher evidence density and less redundant interference under the same budget, enabling stable improvements in fact coverage and generation quality with lower retrieval cost.


%% file: section/6_con.tex
\section{Conclusion}
In this work, we identify two obstacles in knowledge hypergraph generation, poor transfer across scenarios and a structural imbalance where systems over focus on higher order hyperedges.
We introduce Hyper-KGGen, a skill-driven framework that combines coarse-to-fine hyperedge construction with a closed loop skill acquisition mechanism.
By using stability based relative rewards to learn from unstable cases and missing facts, it distills reusable scenario skills and updates a global skill library that improves extraction in new settings.
We also release HyperDocRED, a document level dataset with rigorous $n$-ary hyperedge annotation for evaluating end to end hypergraph generation.
Experiments across datasets and downstream retrieval augmented generation show consistent improvement over strong baselines, indicating that skill evolution and coarse to fine construction paradigm are complementary and jointly improve cross scenario robustness and knowledge hypergraph quality.

\section*{Acknowledgement}
This work was supported by New Generation Artificial Intelligence-National Science and Technology Major Project (No. 2025ZD0124204), the National Natural Science Foundation of China (Nos. U24A20252, U25A20532 and 623B2066), Fundamental and Interdisciplinary Disciplines Breakthrough Plan of the Ministry of Education of China (No. JYB2025XDXM504), and Guangdong Major Project of Basic and Applied Basic Research (No. 2023B0303000009).

%% file: section/7_appendix.tex
\textcolor{white}{ }
\appendix




\section{Baselines}
\label{app:baseline}
In performance comparison, we consider the state-of-the-art baselines, categorized into standard knowledge graph extraction methods, text-based RAG, graph-based RAG approaches, and hypergraph-based RAG models.
\begin{itemize}
    [leftmargin=*,itemsep=0pt,parsep=0.2em,topsep=0.3em,partopsep=0.3em]
    \item \textbf{KGGen}~\cite{mo2025kggen}: an LLM-based text-to-KG generator with iterative entity and edge resolution, reducing sparsity and normalizing relations.
    \item \textbf{RAKG}~\cite{zhang2025rakg}: a document-level KG construction framework that uses pre-entities as retrieval queries to mitigate long-context forgetting and improve global consistency.
    \item \textbf{NativeRAG}~\cite{gao2023retrieval}: a standard vector-based RAG baseline that indexes chunk embeddings in a vector database, and retrieves responses by matching relevant chunks via embedding similarity.
    \item \textbf{GraphRAG}~\cite{edge2024local}: a graph-enhanced RAG that uses LLMs to extract entities and relations into a knowledge graph, and retrieves by traversing the graph structure for global context.
    \item \textbf{LightRAG}~\cite{guo2024lightrag}: a graph-enhanced RAG that fuses structural graph with vector representations, and performs dual-level retrieval over nodes and edges to gather evidence.
    \item \textbf{HiRAG}~\cite{huang2025retrieval}: a hierarchical knowledge RAG framework that leverages the inherent hierarchical knowledge in human cognition.
    \item \textbf{Hyper-RAG}~\cite{feng2025hyper}: a hypergraph-based RAG method that captures higher-order correlations beyond pairwise links, improving semantic faithfulness.
    \item \textbf{HyperGraphRAG}~\cite{luo2025hypergraphrag}: a hypergraph-based RAG framework that represents $n$-ary relational facts via hyperedges and performs retrieval and generation on hypergraph representations.
    \item \textbf{Cog-RAG}~\cite{hu2025cog}: a cognitive-inspired dual-hypergraph RAG approach with theme alignment, performing top-down theme activation followed by entity-level diffusion.
\end{itemize}

\section{Additional Results}
We provide specific results for HyperDocRED dataset in Tab. \ref{tab:micro_macro_prf}.
\begin{table}[htbp]
\centering
\small
\caption{Evaluation of $n$-ary Relations on the HyperDocRED Dataset (Precision, Recall, and F1-score).}
\label{tab:micro_macro_prf}
\renewcommand{\arraystretch}{1.1}
\begin{adjustbox}{max width=0.49\textwidth}
\begin{tabular}{
l
S[table-format=1.4] S[table-format=1.4] S[table-format=1.4]
S[table-format=1.4] S[table-format=1.4] S[table-format=1.4]
}
\toprule
\multirow{2}{*}{\textbf{Method}} &
\multicolumn{3}{c}{\textbf{Micro}} &
\multicolumn{3}{c}{\textbf{Macro}} \\
\cmidrule(lr){2-4}\cmidrule(lr){5-7}
& {\textbf{Pre}} & {\textbf{Rec}} & {\textbf{F1}}
& {\textbf{Pre}} & {\textbf{Rec}} & {\textbf{F1}} \\
\midrule
HyperGraphRAG     & 0.3828 & 0.1072 & 0.1675 & 0.4018 & 0.1208 & 0.1710 \\
HyperRAG          & 0.2439 & 0.2050 & 0.2228 & 0.2486 & 0.2224 & 0.2196 \\
CogRAG            & 0.2884 & 0.1794 & 0.2212 & 0.2926 & 0.1933 & 0.2243 \\
\midrule
Hyper-KGGen              & 0.8327 & 0.3140 & 0.4560 & 0.8436 & 0.3552 & 0.4806 \\
Hyper-KGGen$^+$ & 0.8024 & 0.4300 & 0.5600 & 0.8142 & 0.4609 & 0.5736 \\
\bottomrule
\end{tabular}
\end{adjustbox}
\end{table}

\section{Case Study}
In this section, we present a representative example from the MINE dataset. In this case, Hyper-KGGen achieves 100\% factual coverage under the fact-capture evaluation, substantially outperforming KGGen (40\%) and HyperRAG (46.67\%). Notably, Hyper-KGGen extracts a richer set of entities and relations, exceeding both baselines in volume while still retaining a meaningful subset of binary relations. These results highlight Hyper-KGGen’s strong capability to uncover complex, higher-order relational structures in text.
\subsection{Raw Text}
\begin{myprompt}{The Evolution of Video Games}
    Video games have come a long way since their inception in the mid-20th century. From simple, pixelated games like Pong and Space Invaders to immersive, lifelike worlds in games like Red Dead Redemption and Fortnite, the evolution of video games has been nothing short of extraordinary. This essay will explore the key milestones in the evolution of video games, from the early days of arcade gaming to the rise of console and PC gaming, and the impact of technological advancements on the industry.\newline\newline
    The history of video games can be traced back to the 1950s and 1960s, when computer scientists and engineers began experimenting with interactive electronic games.
\end{myprompt} 

\begin{myprompt}{The Evolution of Video Games}
    The 1970s saw the rise of arcade gaming, with games like Pong, Space Invaders, and Pac-Man becoming cultural phenomena. These games were simple in design but highly addictive, capturing the imagination of players around the world. The success of arcade games paved the way for the development of home gaming consoles, such as the Atari 2600, which brought the arcade experience into people's living rooms.
    The 1980s marked a golden age for video games, with the release of iconic games like Super Mario Bros., The Legend of Zelda, and Tetris. These games introduced new gameplay mechanics, storytelling elements, and visual aesthetics that set the standard for future game development. The introduction of 8-bit and 16-bit graphics allowed for more detailed and colorful game worlds, enhancing the immersive experience for players. \newline\newline
    The 1990s saw the emergence of 3D graphics and CD-ROM technology, leading to the creation of groundbreaking games like Doom, Quake, and Final Fantasy VII. These games pushed the boundaries of what was possible in terms of graphics, gameplay, and storytelling, setting a new benchmark for the industry. The popularity of PC gaming also grew during this decade, with titles like StarCraft, Half-Life, and Diablo becoming instant classics.\newline\newline
    The early 2000s saw the rise of online gaming, with the launch of platforms like Xbox Live and PlayStation Network enabling players to connect and compete with others around the world. Games like World of Warcraft, Halo, and Call of Duty became synonymous with online multiplayer gaming, creating new opportunities for social interaction and collaboration in gaming communities.\newline\newline
    In recent years, the gaming industry has seen a surge in popularity, driven by advancements in technology such as virtual reality (VR), augmented reality (AR), and cloud gaming. VR headsets like the Oculus Rift and PlayStation VR have allowed players to immerse themselves in virtual worlds like never before, while AR games like PokÃ©mon Go have blurred the line between the virtual and physical worlds. The future of video games looks bright, with developments in artificial intelligence (AI), machine learning, and blockchain technology poised to revolutionize the industry. Games are becoming more sophisticated, realistic, and interconnected, offering players new ways to engage with their favorite franchises and immerse themselves in interactive storytelling experiences. In conclusion, the evolution of video games has been a remarkable journey, from the simple black-and-white games of the 1950s to the complex, immersive worlds of today. The industry has grown and evolved in response to advancements in technology, changing consumer preferences, and new opportunities for creativity and innovation. As we look to the future, it is clear that video games will continue to push the boundaries of what is possible, offering new and exciting experiences for players of all ages.
\end{myprompt}

\newpage

\subsection{Knowledge Hypergraph of Hyper-KGGen}
\begin{myprompt_double}{Knowledge Hypergraph generated by Hyper-KGGen}
    \textbf{Entities:}\newline
    video game; pong; space invader; red dead redemption; fortnite; mid-20th century; arcade; console; pc; technological advancement; video game industry; 1950s; 1960s; spacewar!; 1962; massachusetts institute of technology; 1970s; pac-man; atari 2600; 1980s; super mario bros; the legend of zelda; tetris; 8-bit; 16-bit; 1990s; 3d graphic; cd-rom; doom; quake; final fantasy vii; starcraft; half-life; diablo; early 2000s; online gaming; xbox live; playstation network; world of warcraft; halo; call of duty; virtual reality; augmented reality; cloud gaming; vr headset; oculus rift; playstation vr; pokémon go; gaming industry; gaming community; artificial intelligence; machine learning; blockchain technology; industry; player; interactive storytelling experience; future; franchise.

\textbf{Extracted $n$-ary Relations (nodes $\rightarrow$ description)}\newline
\{video game; mid-20th century\} $\rightarrow$ Video games originated during the mid-20th century, establishing the historical starting point for the medium (video game, mid-20th century).\newline
\{video game; pong; space invader; arcade\} $\rightarrow$ Pong and Space Invader serve as early, simple pixelated examples of arcade video games, illustrating foundational forms of the medium (video game, pong, space invader, arcade).\newline
\{video game; red dead redemption; fortnite; technological advancement\} $\rightarrow$ Red Dead Redemption and Fortnite exemplify later, immersive video game experiences that reflect advances in technology (video game, red dead redemption, fortnite, technological advancement).\newline
\{arcade; console; pc; technological advancement; video game industry\} $\rightarrow$ The industry shifted from arcade-focused gaming toward the rise of console and PC platforms, a transition driven and amplified by technological advancement and affecting the broader video game industry (arcade, console, pc, technological advancement, video game industry).\newline
\{1950s; 1960s\} $\rightarrow$ 1950s and 1960s function as the joint formative decades when early interactive electronic game experimentation by computer scientists and engineers originated.\newline
\{spacewar!; 1962; massachusetts institute of technology\} $\rightarrow$ spacewar! is identified with the specific development event that occurred in 1962 at the Massachusetts Institute of Technology, tying the title to a year and institutional origin.\newline
\{1970s; pong; space invader; pac-man\} $\rightarrow$ In the 1970s, Pong, Space Invader, and Pac-Man emerged as prominent arcade titles, indicating these games rose to widespread popularity during that decade.\newline
\{pong; space invader; pac-man; atari 2600\} $\rightarrow$ The commercial success of arcade hits Pong, Space Invader, and Pac-Man helped drive the development of home consoles, exemplified by the Atari 2600, linking those arcade titles to the console's emergence.\newline
\{1980s; super mario bros; the legend of zelda; tetris\} $\rightarrow$ The 1980s is the decade that produced and is characterized by the seminal releases Super Mario Bros., The Legend of Zelda, and Tetris, which came to define that era of gaming.\newline
\{1980s; 8-bit; 16-bit\} $\rightarrow$ The 1980s is associated with the rise of the 8-bit and 16-bit graphical generations, which enabled more detailed and colorful game worlds that shaped games of the era.\newline
\{1990s; 3d graphic; cd-rom; doom; quake; final fantasy vii\} $\rightarrow$ The 1990s saw the emergence of 3D graphics and CD-ROM technology that enabled groundbreaking titles such as Doom, Quake, and Final Fantasy VII to push audiovisual and design boundaries.\newline
\{1990s; pc; starcraft; half-life; diablo\} $\rightarrow$ In the 1990s, PC gaming grew in popularity and was marked by defining PC titles such as StarCraft, Half-Life, and Diablo that became instant classics on the platform.\newline
\{early 2000s; online gaming; xbox live; playstation network\} $\rightarrow$ In the early 2000s, the expansion of online gaming was driven by platform launches—Xbox Live and PlayStation Network—that enabled large-scale player connectivity.\newline
\{world of warcraft; halo; call of duty; online gaming; gaming community\} $\rightarrow$ World of Warcraft, Halo, and Call of Duty acted as flagship online multiplayer titles that shaped online gaming practices and fostered social interaction and collaboration within gaming communities.\newline
\{gaming industry; virtual reality; augmented reality; cloud gaming\} $\rightarrow$ Recent growth in the gaming industry has been driven by technological advances—virtual reality, augmented reality, and cloud gaming—that expanded how games are experienced and who can access them.\newline
\{augmented reality; pokémon go\} $\rightarrow$ Pokémon Go served as a prominent augmented reality game that merged digital gameplay with physical-world interaction, blurring boundaries between virtual and real environments.


\{artificial intelligence; machine learning; blockchain technology; industry; video game; future\} $\rightarrow$ Developments in artificial intelligence, machine learning, and blockchain technology are expected to transform the video game industry and shape the future direction of video games.\newline
\{video game; player; franchise; interactive storytelling experience\} $\rightarrow$ Modern video games provide players new ways to engage with franchises and to be immersed in interactive storytelling experiences.\newline
\{industry; video game; player; future\} $\rightarrow$ The industry has grown and transformed over time, and, looking toward the future, will continue pushing boundaries through video games to deliver new experiences for players.
\end{myprompt_double}